\documentclass[runningheads]{llncs}
\usepackage{graphicx}
\usepackage{tikz}
\usepackage{comment}
\usepackage{amsmath,amssymb} 
\usepackage{color}
\usepackage{cite}
\usepackage{tabularray,booktabs}
\usepackage{multirow}
\usepackage{diagbox}
\usepackage{adjustbox}
\usepackage{caption}
\usepackage[labelformat=simple]{subfig}
\usepackage[utf8]{inputenc}
\usepackage{amsmath}

\begin{document}
\pagestyle{headings}
\mainmatter
\def\ECCVSubNumber{5935}  

\title{Biasing Like Human: A Cognitive Bias Framework for Scene Graph
  Generation} 

\titlerunning{ A Cognitive Bias Framework for Scene Graph
Generation}
%
\author{Xiaoguang Chang \and
  Teng Wang \and
  Changyin Sun \and
  Wenzhe Cai}
\authorrunning{Xiaoguang Chang et al.}
%
\institute{
  School of Automation, Southeast University, China\\
  \email{\{xg\_chang,wangteng,cysun,wz\_cai\}@seu.edu.cn}\\
}
\maketitle
\begin{abstract}
  Scene graph generation is a sophisticated task because there is no specific
  recognition pattern (e.g., \textit{looking at} and \textit{near} have no
  conspicuous difference concerning vision, whereas \textit{near} could occur
  between entities with different morphology ). Thus some scene graph generation
  methods are trapped into most frequent relation predictions caused
  by capricious visual features and trivial dataset annotations. Therefore,
  recent works emphasized the “unbiased” approaches to balance predictions for a
  more informative scene graph. However, human's quick and accurate judgments
  over relations between numerous objects should	be attributed to \textbf{bias}
  (i.e., experience and linguistic knowledge) rather than pure vision. To enhance
  the model capability, inspired by the \textbf{“cognitive bias”} mechanism,  we
  propose a novel 3-paradigms framework that simulates how humans incorporate the
  label linguistic features as guidance of vision-based representations to better
  mine hidden relation patterns and alleviate noisy visual propagation.  Our
  framework is model-agnostic to any scene graph model. Comprehensive experiments
  prove our framework outperforms baseline modules in several metrics with
  minimum parameters increment and achieves new SOTA performance on Visual Genome
  dataset.

\end{abstract}

\section{Introduction}
Scene graph generation refers to the vision task that detects objects and
recognizes semantic relationships between different objects in an image. With
graph-based representation, we could depict the semantic content of images in a
structural and efficient way. Such a structural representation of images could
greatly benefit downstream vision tasks, including image
captioning\cite{ic1}\cite{ic3}, visual question answering
\cite{vqa3}\cite{vqa2}\cite{vqa1}, and video understanding\cite{va1}.

\begin{figure}[t!]
  \subfloat[]{\includegraphics[width=0.3\textwidth,,height=3cm]{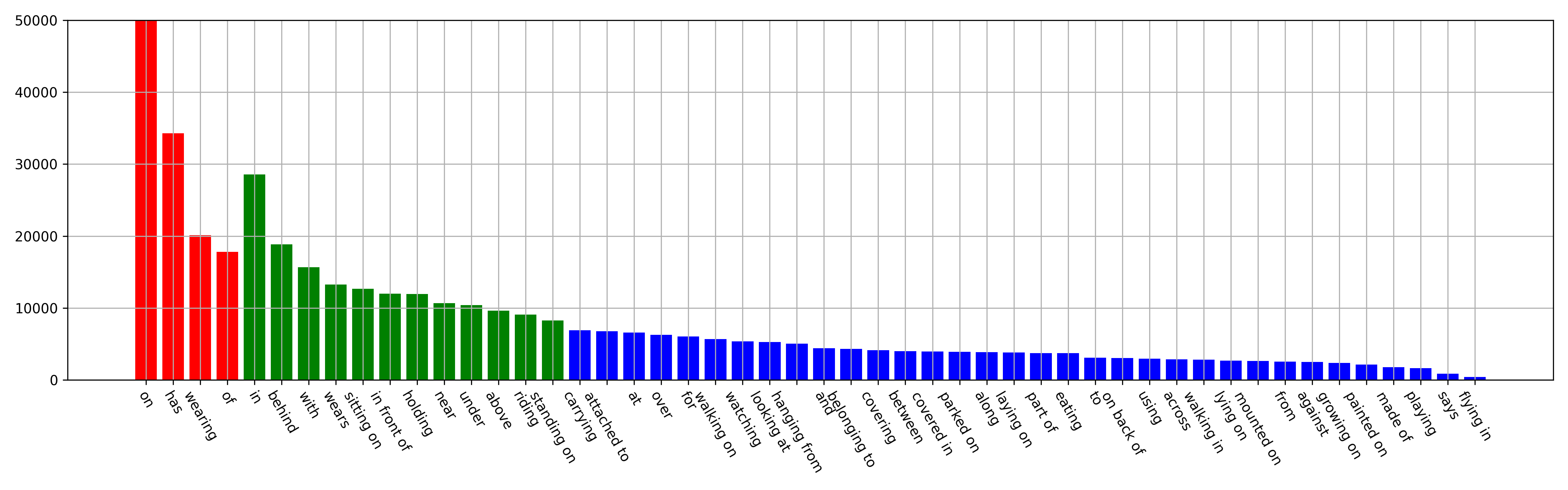}\label{sub:1}}\quad
   \subfloat[]{\includegraphics[width=0.3\textwidth,height=3cm]{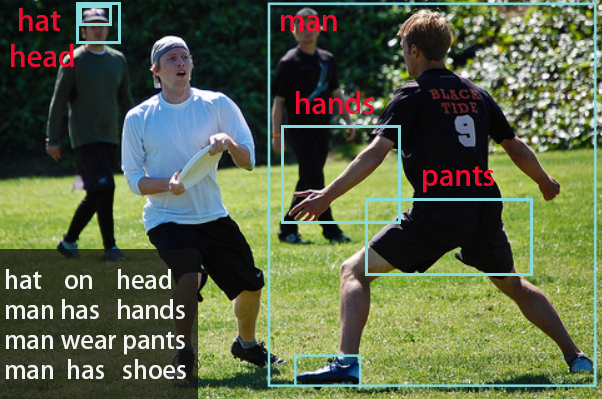}\label{sub:2}}\quad
  \subfloat[]{\includegraphics[width=0.3\textwidth,height=3cm]{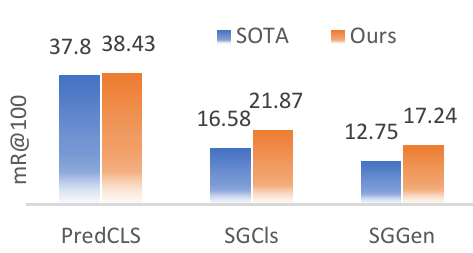}\label{sub:3}}
  \caption{\protect\subref*{sub:1}: predicate sampling frequency in VG dataset. \protect\subref*{sub:2}: {an annotation example on VG dataset with numerous tedious relations }. \protect\subref*{sub:3}: Mean recall@100 improvement of our framework on three tasks over SOTA.}
  \label{intro}
  
\end{figure}

However, the application of scene graph is confined by full of low semantic
predictions in previous scene graph models. The inferior predictions is a
consequence of several factors. First, the relationship sample distributions in
existing representative datasets are long-tailed. For example, the sum of
relations \textit{on},\textit{in}, \textit{of} in the widely adopted Visual
Genome dataset\cite{visualgenome} counts for more than 50\% of the total
(Figure.\ref{sub:1}).  Therefore the network shows a strong tendency to head
relations and behaves poor on most tail classes. Besides, the image annotation
is incomplete. Those informative but distant relations are typically ignored
and thus falsely treated as negative samples during learning. As shown in
Figure\ref{sub:2}, Second, those tail relations (i.e., \textit{play, see})
typically involve complex semantic meaning and large intra-class variations in
images. Large intra-class variations combined with few samples make feature
representation learning hard for those tail relations.

To address the aforementioned challenges, a large number of studies have been
conducted to enhance quality of predictions. Concerning long-tailed data
distribution issue,  recent works strive for designing ``debiased'' methods,
which re-design inference procedures or loss functions to obtain a balance
prediction. Nonetheless, these ``debiased'' methods could be seen as a special
form of ``re-weighting'' but in a learnable way. As no extra feature
enhancement modules are introduced, these methods improve predictions on tail
categories at the expense of those head categories. As for better feature
representation learning, different propagation mechanisms have been developed
for modeling context\cite{bgnn}. Nevertheless, it fails to achieve satisfactory
performances due to noisy propagation.

\begin{figure}[t!]
  \centering
  {
    \includegraphics[width=0.95\textwidth]{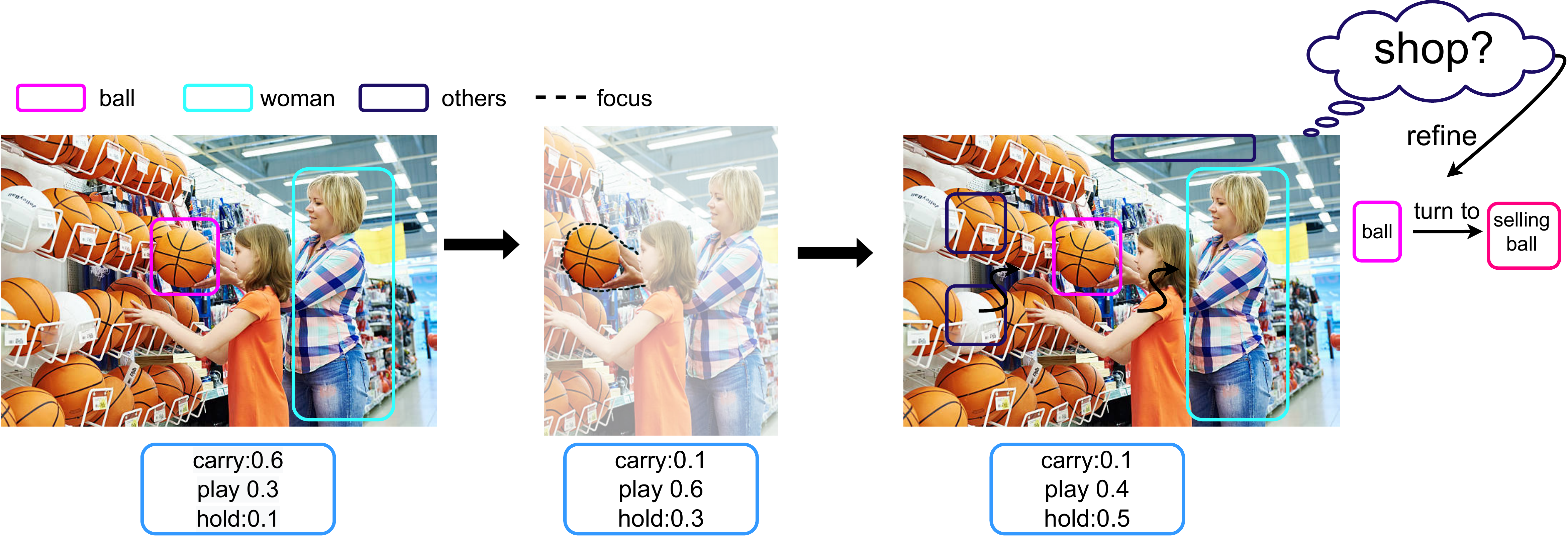}
  }

  \caption{An example of three human Cognitive paradigms. (Left): Given two
    categories ``woman'' and ``ball'', the relation distribution is generated by
    the experience of combinations. (Middle): Considering two objects properties
    (one belongs to goods, another is human-being ), intersect region (i.e. hands)
    indicates that their relation shell be	\textit{hold, play} rather than
    \textit{carry}. (Right): For further determination, Global contexts are
    incorporated into a single object description by prefixing ``selling'' to
    \textit{ball}, which provides a tendency to prefer \textit{hold}.}
  \label{fig:para}

\end{figure}

However,  bias should not be alleviated arbitrarily. In human recognition
pipeline, \textbf{``Cognitive bias''}\cite{cogbias}, as  a psychological
activity,  benefits our decision-making in uncertain and intricate
circumstances with the help of language\cite{advbias}. Figure\ref{fig:para}
illustrates how  cognitive bias impacts on relation cognition. Given a ball and
woman entities,
human first exploits experience of several  most frequent occurrence relations
between ``ball'' and ``woman'', these candidates are served as the alternative
choices for later decisions . Next, due to the experience can not represent the
current circumstance, more precise guesses can be obtained by focusing on the
part that woman contact the ball, which is the typical criteria when determines
the relation between a person and a object. Noticed that the criteria is not
necessary to be intersection part of two entities, but affected by categories.
In the end, by considering all objects in this scene, it is not hard to
conclude a shopping scenario. Hence human can choose ``hold'' as best choice by
alleviating univocal meaning of ``ball'' with restriction ``selling''.
In this work, inspired by three \textbf{cognitive bias} paradigms, we propose a
language-based framework to enhance feature representations and reduce visual
feature redundancy , by extracting  prior knowledge of experience, linguistic
local and global context respectively.	Concretely,  We first utilize object
pair labels to simulate the experience over predicates to same space of
prediction likelihoods in a supervised manner. Then, a language map module
projects linguistic similarity as a channel-wised attention which focusing on
local specific visual patterns that can impact final decisions. Moreover, a
scene extractor module refines each object’s linguistic representation by the
global context.   To our best knowledge, we are the first to introduce human
\textbf{cognitive bias} in Scene graph generation task, and utilize labels as
high dimension representations. We conduct widely and detailed experiments on
Visual Genome Dataset to prove framework effectiveness. The main contributions
include:
\begin{enumerate}
  \item We propose a 3-paradigms cognitive bias framework that	simulates human
        thinking procedure to effectively  extracting informative relation patterns
        from images and remedy the trivial dataset problem.
  \item we design 3 cognition-driven language feature representations as the
        guidance of visual-based message propagation in  the form of bias.
  \item Our proposed framework is flexible and lightweight that can be plug-in
        any out-of-shelf scene graph model. And we achieve new SOTA performance in
        three scene graph tasks on baseline BGNN\cite{bgnn} with margins.
        (Figure\ref{sub:2})

\end{enumerate}
\section{Related Work}
\textbf{Scene Graph Generation:} In the community, there are mainly three
mainstream methods for scene graph generation. Early works widely adapted
message passing between all object
proposals\cite{drnet}\cite{Gps-net}\cite{mem}\cite{linknet}\cite{factorizable}\cite{structure},
or aggregating within different type nodes\cite{phrases}. This method commonly
focused on the global context. A number of them\cite{motif}\cite{phrases}
harnessed sequential memory model(LSTM\cite{lstm} or GRU\cite{gru}), but failed
to adopt edge feature in context formulation. Later
works\cite{bgnn}\cite{energy}\cite{grcnn} harness Graph neural
network(GCN\cite{gcn}) processing node and edge features respectively, and
focusing more on pair-wise information.\cite{bgnn} applied a multi-stage graph
message propagation between proposal entities and relationship representation.
In \cite{grcnn}, Yang \textit{et al.} pruned graph connections to sparse one,
then attentional graph convolution network is applied for modulating
information flow. \cite{energy} utilized GCN for updating state representations
as energy value.  Chen \textit{et al.}\cite{kern} constructed a graph between
proposal and all relationship representations and aggregated messages by GRU.
Yet, those works suffered from the drawback that GCN only propagates features
on local context deteriorates model to draw dominant predicates, which was
addressed by\cite{unbias} as biased training. Aiming to fix this shortcoming,
recent works proposed novel pipeline and utilized prior knowledge
\cite{unbias}\cite{fully}\cite{pcpl}\cite{unconditional}\cite{recovering}\cite{attention-translation}.
serving as a plugin to generalized scene graph model.

\noindent\textbf{Prior Knowledge:} Current works addressed the fundamental role
of prior knowledge in terms of the long-tail problem.
The commonly adopted method is leveraging statistical results\cite{motif} as
addictions on the decision layer. For instance, ``FREQ''\cite{motif} directly
summed prior distribution to model decision layer by counting
``subject-relation-object'' co-occurrence in dataset. However, the counting
approach is hard to mimic real word distribution because of low-quality
annotations in datasets. Therefore, an increased number of works strived for
leveraging prior knowledge in perception procedure. \cite{lp} designed a
language module, which concatenated two retrained word vectors and projected
representation to the same space of vision module for minimizing the distance
of similar semantic features. \cite{schemata} entangled the perception and
prior in a single model with shared parameters trained by multi-task learning.
\cite{vctree} introduced a confidence estimation module to alleviate the error
propagation by incorporating confidence estimation in graph node feature
updating. \cite{bridge} incorporated external commonsense knowledge by unifying
the formulation of scene graph and commonsense graph
\cite{kern} incorporates co-occurrence of objects and relations to form a
knowledge graph as a constraint in graph message propagation.
\cite{commonsense} structured visual commonsense and proposed a cascaded fusion
architecture for fusion. Our model combines both the decision layer and the
feature processing layer's fusion.

\noindent\textbf{Scene Graph Debasing:} Due to the lang-tail problems, some
researchers proposed the "unbiased" concept, which prevents model overfitting
to most frequent categories. Early works simply adopted weighted loss or focal
loss\cite{focal} for balanced prediction. Later, Yan \textit{et al.}\cite{pcpl}
designed a training scheme that loss weight is dynamically changed based on
predicate's visual feature similarity, which is learned by minimizing the
distance of similar predicate features. Nonetheless, vision patterns of
relations are highly fickle, hence hard to learn.  \cite{unbias} presented a
general unbiased inference pipeline that utilized Total Direct Effect analysis.
However, this approach did not enrich model capability, but also wiped out
object labels information, which can bring out beneficial language cues.  On
the contrary, our approach exhaustively uses labels for assisting vision.
Experiments in \ref{Quantitative} support our hypothesis.

\begin{figure}[t!]
  \subfloat[Overall
    Architecture]{\includegraphics[width=0.98\textwidth]{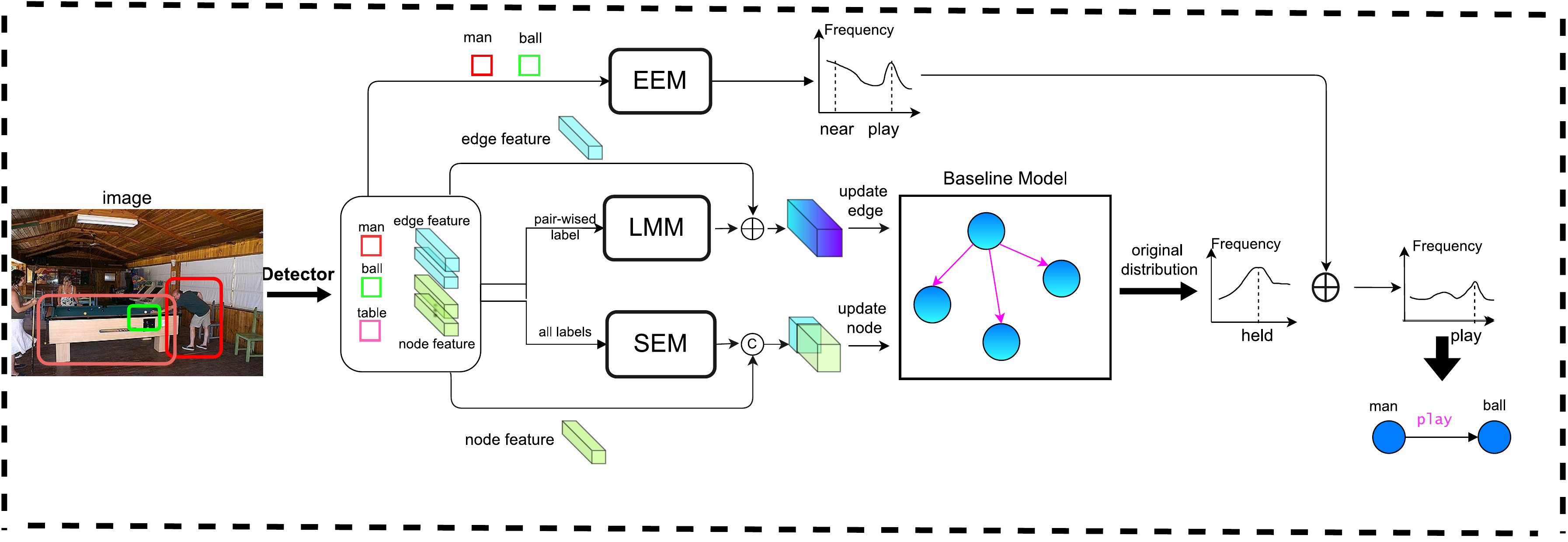}\label{sub:overall}}\\
  \subfloat[Linguistic map
    module]{\includegraphics[width=0.48\textwidth]{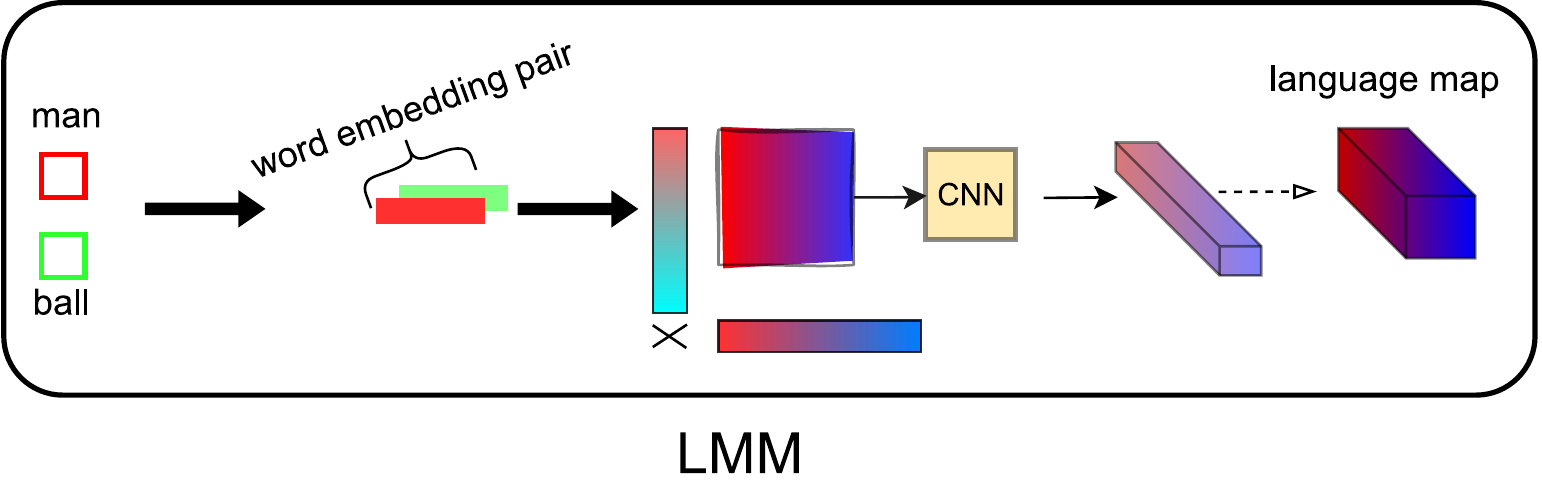}\label{sub:lmm}}\quad
  \subfloat[Scene extractor
    module]{\includegraphics[width=0.48\textwidth]{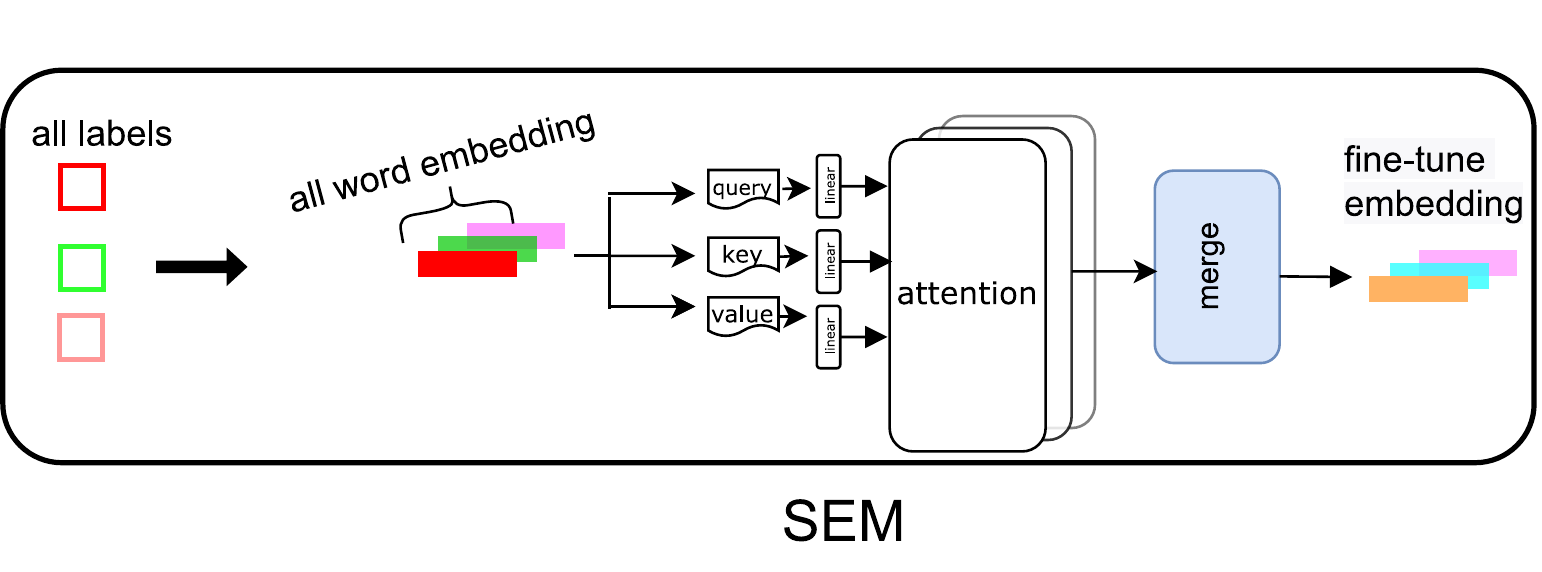}\label{sub:sem}}
  \caption{Overview of our proposed C-bias framework. First, we get
    predicted object labels and consist object and edge features from ROI features.
    Than \textit{Linguistic map module (LMM)} receives object pairs label to
    generate a local language map that as a channel bias on edge features. Next,
    \textit{Scene extractor module (SEM)} takes all labels in global context into
    consideration and generates divergent linguistic representations to update node
    features. The node and edge representations are updated before baseline model.
    Finally, \textit{Experience Estimation Module (EEM)} yields estimated relation
    distribution by label pairs, and updates the final relation likelihood.}
  \label{net}
\end{figure}

\section{Methodology}
\subsection{Problem Formulation}
Given an image $I$, scene graph generation task aims to predict object
coordinates (bounding box) $B \in \mathbb{R}^{4}$ , classes $C \in \mathbb{R}$
as well as relation $R$ between objects. The task can further denote as the
optimization of $P\left(R,B,C\mid I\right)$. For a graph structure, it
normally consists of nodes and edges. We denote $n_{i} \in {\mathcal{V}}$ as
i-th object representation. $e_{i, j} \in {\mathcal{E}}$ as edge representation
from node $i$ to $j$,

\subsection{Framework Architecture}
Our proposed cognition bias (C-bias) framework overall architecture is
presented in Figure\ref{sub:overall}, which is consist of three paradigms. 1)
\textit{the experience estimation module} is to draw the relation distribution
on supervision of joint possibility. This distribution is added on final model
outputs. 2) \textit{the language map module} (Figure\ref{sub:lmm}) is to learn
local interactions between pair labels and produce channel-wised attention for
visual features. 3) \textit{the scene extractor module} (Figure\ref{sub:sem})
is to globally consider the linguistic context in an image, and then update the
initial label embedding. Those three sub-modules construct an enhanced graph
representation by introducing label features in the vision-based baseline model
at the output, edge, node, respectively.
\setcounter{secnumdepth}{3}
\subsubsection{Object Detection Network:}~\\
Concretely, given an image, a pre-trained object detection network is used to
obtain a predicted object location set $\hat{{B}}= \lbrace
  b_{i}\rbrace^{n}_{i=1}$ and a class set $\hat{{C}}= \lbrace
  c_{i}\rbrace^{n}_{i=1}$ with number $n$.  Detection network can be written as:
\begin{equation}
  P(\hat{B},\hat{C}\mid I)=\operatorname{detector}(I)
\end{equation}
where $I$ is an input image.
\subsubsection{Paradigm 1: Experience Estimation Module:}~\\
This module learns the joint likelihood of two objects over relationships. We
expect it can output distributions that base on the experience of correlation
between predicates and categories. The estimated experience can be acquired by
MLP layers, with the input of pair label categories $c_{i}, c_{j}$ and position
embedding $p_{ij}$. Here, we use different weight $w_{s}$ and $w_{o}$ to
transform subject and object labels to their linguistic representations,
respectively. The predicted distribution between $i$-th subject and $j$-th
object $d_{ij}\subseteq\mathbb{R}^{N_{r}}$ can be describe as:
\begin{align}
  \centering
  p_{ij} & =\phi_{p}(x_{i},y_{i},w_{i},h_{i},x_{j},y_{j},w_{j},h_{j})\label{pos
  encoding},                                                                    \\
  d_{ij} & =\varphi\left[\phi_{s}(w^{T}_{s}c_{i}):
  \phi_{o}(w^{T}_{o}c_{j}):p_{ij}\right]\label{est}.
\end{align}
Where $x,y,w,h$ are object coordination, width, height.
$\phi_{p},\phi_{s},\phi_{o},\varphi$ are fully connect layer with activation
function (e.g. RELU). $[:]$ is concatnation operation.

Here, unlike previous ``FREQ''\cite{motif} method that directly counts triblets
$\left\langle sub,rel,obj \right\rangle$ co-occurrence frequency as prediction
likelihood weights, we utilize the joint possibility as supervision signals to
obtain a comprehensive and accurate relation distribution over tail categories.
Suppose the dataset contains $N_{o}$ number of object categories and  $N_{r}$
relation categories, the joint possibility of subject i with object j is
denoted as:
\begin{equation}
  P_{rel}^{ij} =s_{i}\times o_{j}.
\end{equation}

\begin{figure}[tbp!]
  \centering
  \begin{minipage}[t]{0.48\textwidth}
    \centering
    \subfloat[man: subject]
    {

      \includegraphics[width=5.7cm,height=3cm]{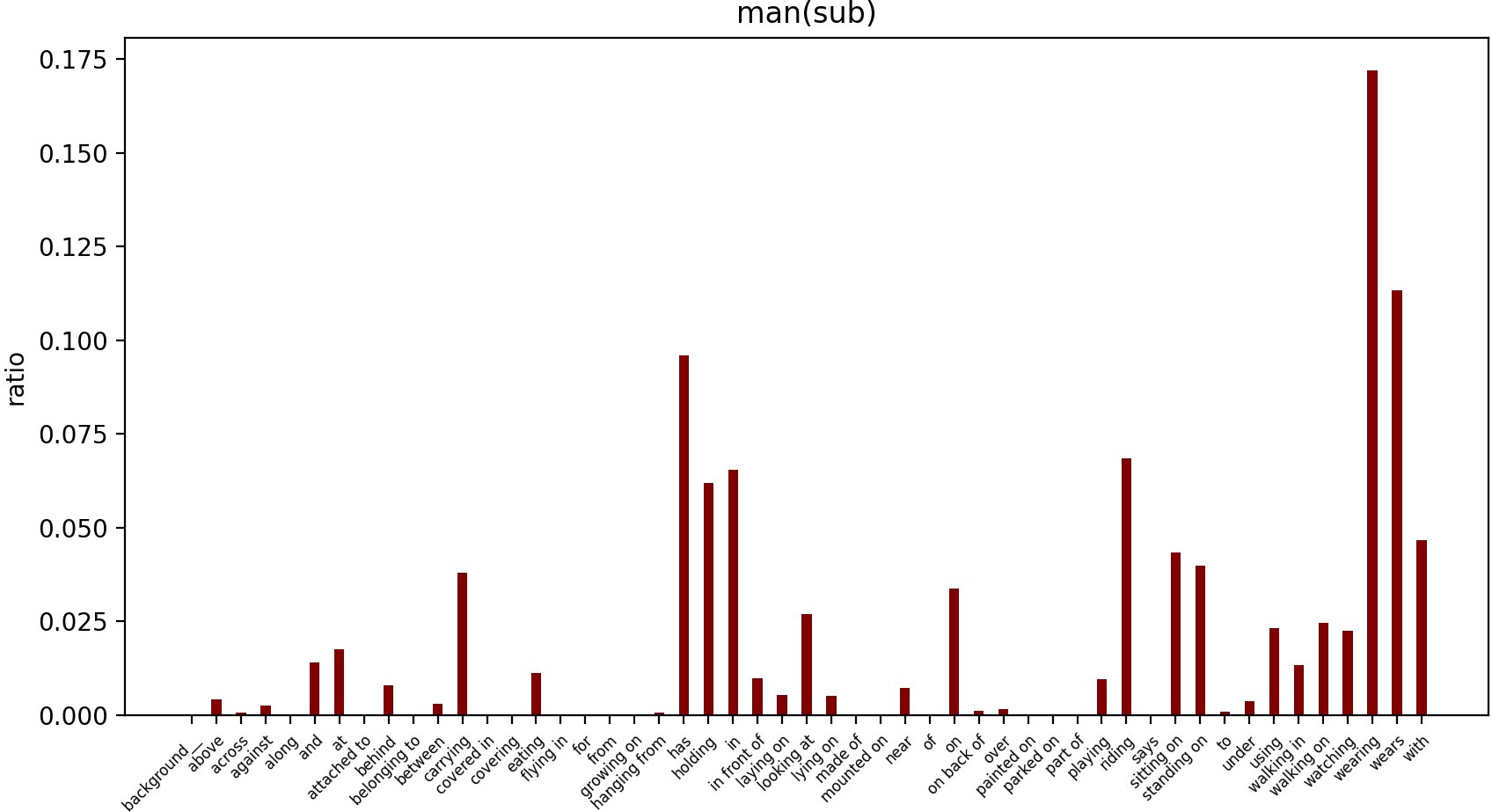}
      \label{sub:man}

    }\qquad
  \end{minipage}
  \begin{minipage}[t]{0.48\textwidth}
    \centering
    \subfloat[man: object]
    {

      \includegraphics[width=5.7cm,height=3cm]{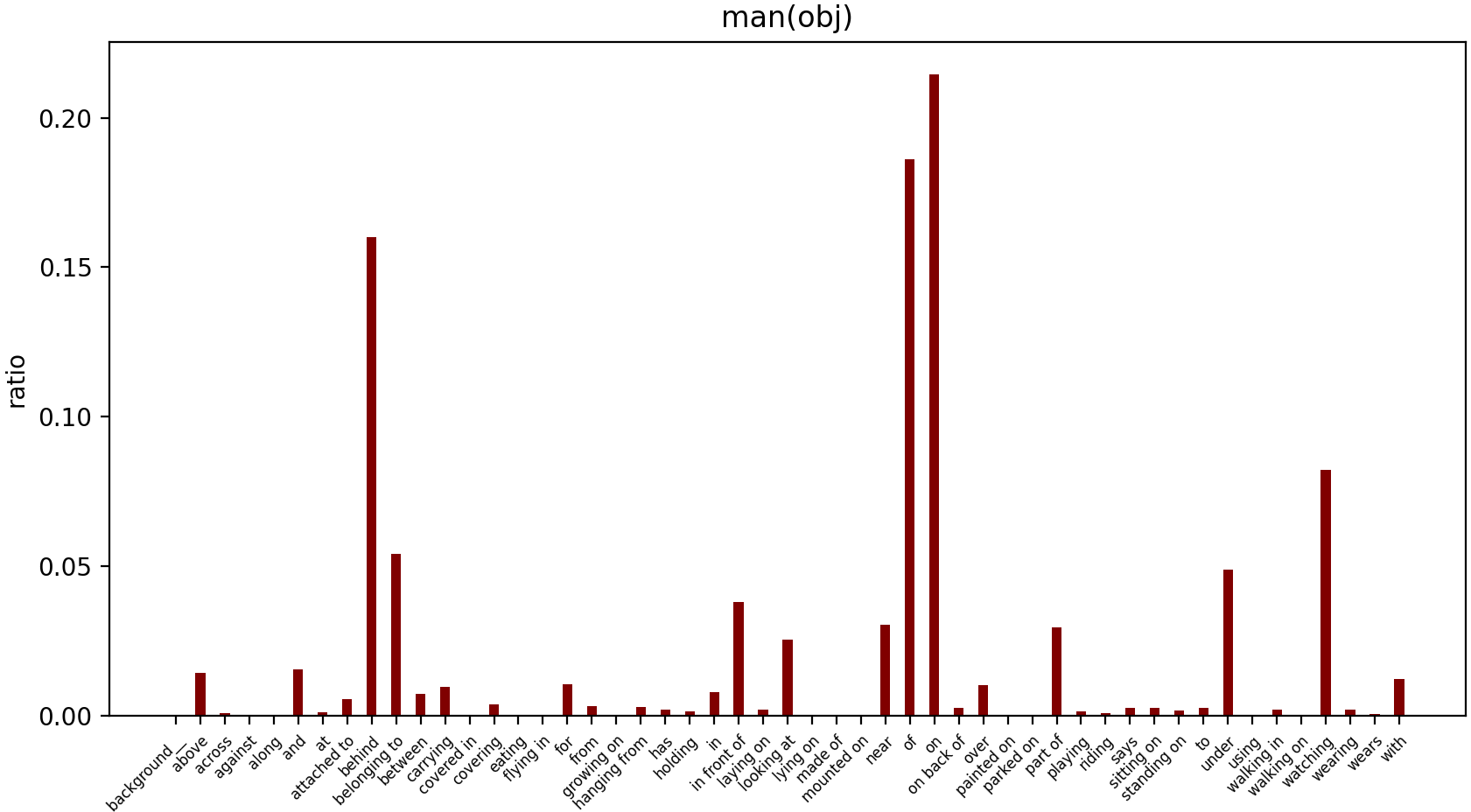}
      \label{obj:man}

    }
  \end{minipage}
  \caption{Distribution discrepancy between subject and object.}
  \label{fig:Distribution Discrepancy}
\end{figure}
Where $s_{i},o_{j}\subseteq\mathbb{R}^{N_{r}} $ are statistical  marginal
distribution  of subject and object categories over relation categories (e.g.,
Figure\ref{fig:Distribution Discrepancy}). Compare to directly counting
$N^{2}_{o}$ number of triblet combinations, marginal distribution does not
suffer a lot from sample scarcity, thus can generate more accurate
distribution. Noticed that the relation distribution of the same object classes
is alternated depending on whether it serves as a subject or object.
For instance, Figure\ref{sub:man} shows that ``man'' as the subject is possible
having correlation with ``eating''. Inversely, in Figure\ref{obj:man}, it is
impossible.

Finally,  For learning distribution purpose, we choose cosine loss function
between target $P_{rel}^{ij}$ and predicted prediction $d_{ij}$, which is
denoted as:
\begin{equation}
  \mathcal{L}_{est}=\sum_{i,j}^{N}1-\cos(d_{ij},P_{rel}^{ij}).
  \label{lossest}
\end{equation}


\subsubsection{Paradigm 2: Language Map  Module:}~\\

To incorporate	the language feature with local linguistic context into visual
feature, we design a ``language mapping'' (LM) operation to generate linguistic
attention based on label pairs $c_{i}, c_{j}$ and project it to the same
dimensions as the edge (conventionally, it is the visual feature of union
area). Say we got two predicted class labels $c_{i}$ and $c_{j}$,
the initial language map $x_{ij}$ is produced by the vector multiplication as
follows:
\begin{equation}
  \label{in}x_{ij}=\left< (w^{T}_{s}c_{i})^T, w_{o}^{T}c_{j} \right>,
\end{equation}
Where $w_{s}$, ${w_{o}\in \mathbb{R}^{D_{w}}}$ are word embedding
weights of subjects and objects. $\left<,\right> $ is dot product operation.
The language map initially can be regarded as a covariance matrix between
subject and object word embeddings, which contains informative correlation
information.
Then, pooling and 2D-convolution are applied to gain channel-wised linguistic
attention. The language map feature $f_{LM}\in \mathbb{R}^{C \times 1\times 1}$
can be denoted as  Eq.\ref{lm}.


\begin{equation}
  f_{LM}=\sigma (G^{N}_{\operatorname{conv}}...\sigma
  (G^{1}_{\operatorname{conv}}(G_{\operatorname{pooling}}(x_{ij})))).
  \label{lm}
\end{equation}
Where N is number of convlution 2D layers, $G_{\operatorname{pooling}}$ can be
Avgpooling, $\sigma$ is RELU function.
Finally, Eq.\ref{language map}	is applied to get final language attention bias
$l_{ij}$.
\begin{align}
  \label{language map} & l_{ij}=Upsampling(f_{LM}\left(x_{ij}\right)),
  \\
  \label{edge}         & \hat{e}_{ij}= e_{ij}+l_{ij}.
\end{align}
Where generated $l_{ij} \in \mathbb{R}^{C \times D_{p}\times D_{p}}$ is as the
same dimension $D_{p}$ as ROIPooling  feature $e_{ij} $. Next, the original
edge feature ${e}_{ij}$ is updated in Eq.\ref{edge}, as an item of bias.

\subsubsection{Paradigm 3: Scene Extractor Module}~\\

The aforementioned paradigms both generate deterministic linguistic
representations given the input label pair. However, the same word in different
contexts could have different semantic meanings. Hence, the purpose of this
module is to extract global semantics context into each category
representation. Hence we leverage multi-head self-attention to achieve that.
Differing from prior work\cite{seq2seq} taking visual object features as
transformer input, this module only takes object labels concatenated with
positions for context encoding.  Specifically, our module consists of several
stacked multi-head attention layers with a Feed Forward network (FF) to update
each node (object) feature. Given an image, suppose there are $ {n}$ object
proposals that are packed together as input $ {I}$.

\begin{equation}
  {I}=\left\{ \left[ {w}^{T}_{c}c_{i}:p_{i}\right] \right\}_{i}^{n}.
  \label{att_input}
\end{equation}
Where $w_{c}$ is object's label embedding weights  and $p_{i}$ is position
embedding of i-th proposal as the same formulation as Eq.\ref{pos encoding}. $
  \left[ : \right]$ is concatenation operation. Then attention mechanism can be
written as:
\begin{equation}
  \operatorname{Attention}\left(Q,K,V\right)=\operatorname{softmax}\left(\frac{Q
    K^{T}}{\sqrt{d_{k}}}\right) V.
  \label{attention}
\end{equation}
Where $Q=I W^{Q}, K=I W^{K}, V=I W^{V}$ refer to query, key, value
respectively, $W^{Q}, W^{K},W^{V}$ corresponding to parameter matrices, $d_{k}$
is the dimension of key. So we formulate multi-head attention as:

\begin{align}
  \operatorname{Multihead}\left(Q,K,V\right) & =\psi\left(
  head_{0},head_{1}...head_{n} \right),                                                     \\
  \text{where}
  \operatorname{head}_{i}                    & =\operatorname{Attention}\left(Q,K,V\right).
  \label{SCENE}
\end{align}
Where $\psi$ is a MLP layer with the purpose of merging all attention heads.
Finally, scene representation of i-th object $s_{i}$  can be described as:

\begin{align}
  \operatorname{FFN}(x) & =\max \left(0, x W_{1}+b_{1}\right) W_{2}+b_{2},    \\
  s_{i}                 & =\operatorname{LayerNorm}(x+\operatorname{FFN}(x)).
  \label{fpn}
\end{align}
Where $x$ is output of multi-head attention. $W_{1},W_{2},b_{1},b_{2}$ are FPN
parameter matrices and biases.
Then the $i$-th node feature ${n_{i}}$ is updated:

\begin{equation}
  \hat{n_{i}}=s_{i}\oplus n_{i}.
  \label{multi-head}
\end{equation}
Where $\oplus$ is concatenation operator.
\subsubsection{Baseline Model}~\\
After edge and object features are updated, any off-the-shelf scene graph
generation network can be used. It could be either a message propagating method
or a graph neural network, as long as it exerts node or edge features for
prediction. This network can be described as:
\begin{equation}
  P(R|I,B,C)=\operatorname{SGG}(\hat{N},\hat{E}).
  \label{sgg}
\end{equation}
Where $I,B,C$ is a given image, Object location set, class set.
$\hat{N}=\left\{ \hat{s_{i}}\right\}$ and $\hat{E}=\left\{
  \hat{e_{ij}}\right\}$ are node and edge features enhanced by language. Then,
SGG model output and "Paradigm 1" estimated relation distribution $D=\left\{
  {d_{ij}}\right\}$ are collectively considered:
\begin{equation}
  \hat{P}(R|I)=\operatorname{SGG}(\hat{N},\hat{E})+D.
\end{equation}
\section{Experiments}
\label{experiments}
In this section, we first introduce the dataset and evaluation metrics we used
as well as the implementation details. Then the comprehensive comparison
experiments with other scene graph methods and ablation studies on three
paradigms are conducted. Finally, we demonstrate quantitative results for
illustration purposes.
\subsection{Evaluation Protocol}
\textbf{Datasets}:  We train and evaluate on filtered Visual Genome split VG150
proposed in\cite{imp} to keep consistency with other studies. The VG150 dataset
contains the most frequent 150 objects and 50 relationships with 108k images,
in which 70\% images are held out for training and 30\% for testing. Among the
training set, 5000 images are fetched for evaluation.\\

\noindent\textbf{Tasks:} 1) \textit{predicate classification (PredCls)}  task
predicts predicates given the ground truth locations and classes. 2)
\textit{scene graph classification (SGCls)} task recognizes predicates and
classes given locations.  \textit{scene graph generation (SGGen)} predicts
valid objects (IoU $>$ 0.5) and classes as well as predicates.\\

\noindent\textbf{Metrics:} Following the convention of the previous studies, We
take mean  recall@k, Zero-Shot Recall\cite{lp}, and No Graph Constraint mean
Recall\cite{pixels} as evaluation metrics. We do not report
recall@k\cite{vctree} because it mainly focuses on low-semantic relations.
\subsection{Implementation Details}
All models are training on two NVIDIA Titan XP GPUs. The learning scheduler is
WarmupReduceLROnPlateau with decay factor 0.6 and patient 6. The base learning
rate is 0.08,0.08,0.06, and batch size is 12. We choose the SGD optimizer for
optimization.\\

\noindent\textbf{Detector:} We use pretrained faster-rcnn\cite{rcnn}  with
backbone ResNeXt-101-FPN\cite{maskrcnn} . We froze its weights during training.
\\

\noindent\textbf{C-bias Framework:} We test BGNN\cite{bgnn}, Motifs\cite{motif}
and G-RCNN\cite{grcnn} as baseline. For a fair comparison, all configurations
in our proposed network will be identical to baseline settings, including node
and edge feature dimensions. Noticed that for each baseline model, we choose
the same configurations for all 3 tasks. We use pre-trained
glove.6B\cite{glove} with $D_{w}=200$ as the initial word embedding weights
$W_{s}$,$W_{o}$ for three paradigms.
Specifically, for paradigm 1, we use 3 FC layers $\Phi_{s},\Phi_{o},\Phi_{p}$
with 1024 neurons and $\varphi$ is a 2 layers MLP with hidden dimension 4096.
For paradigm 2, We choose two $3\times3$ convolution layers to generate a
256-channels language feature map. The output size $D_{p}$ of
ROIAlign\cite{maskrcnn} is 7$\times$7. For paradigm 3, We use 4 layers
self-attention with 8 heads, of which output 512-dim linguistic-based object
features $s$.

\begin{table}[!t]
  \centering
  \caption{The SGG performances of constraint and no graph constraint mean
    Recall@K  in \% on VG dataset. * denotes results come from author's paper. /
    means results are not available}
  \label{tab:eval}
  \resizebox{\linewidth}{!}{%
    \begin{tabular}{lccccccccccc}
      \hline
                                                  & \multicolumn{1}{l}{\multirow{2}{*}{Model}} &
      \multicolumn{1}{l}{\multirow{2}{*}{Method}} & \multicolumn{3}{c}{PredCls}
                                                  & \multicolumn{3}{c}{SGCls}
                                                  & \multicolumn{3}{c}{SGGen}
      \\
                                                  & \multicolumn{1}{l}{}                       &
      \multicolumn{1}{l}{}                        & ~ ~mR@20                                   & mR@50
                                                  & mR@100~ ~                                  & ~ ~mR@20               & mR@50          &
      mR@100~ ~                                   & ~ ~mR@20                                   & mR@50                  & mR@100~ ~
      \\
      \hline
      \multirow{9}{*}{constraint}                 & PCPL*\cite{pcpl}
                                                  & /                                          & /                      & 35.2
                                                  & 37.8                                       & /                      & 18.6
                                                  & 19.6                                       & /                      & 9.5            &
      11.7                                                                                                                                 \\
                                                  & IMP*\cite{imp}
                                                  & /                                          & /                      & 15.8
                                                  & 17.2                                       & /                      & ~9.3
                                                  & 9.6                                        & /                      & 6.0            &
      7.3                                                                                                                                  \\
                                                  & VCTree*\cite{vctree}
                                                  & /                                          & /                      & 15.4
                                                  & 16.6                                       & /                      & ~ ~7.4~ ~
                                                  & 7.9                                        & /                      & 8.2
                                                  & 9.7                                                                                    \\
      \cline{2-12}
                                                  & \multirow{2}{*}{G-RCNN\cite{grcnn}}
                                                  & \multicolumn{1}{l}{baseline}               & 13.21                  & 16.46
                                                  & 17.28                                      & 8.35                   & 9.67
                                                  & 10.17                                      & 3.45                   & 4.89           &
      5.95                                                                                                                                 \\
                                                  &                                            &
      C-bias                                      & \textbf{15.75}                             & \textbf{18.64}
                                                  & \textbf{19.80}                             & \textbf{8.62}          & \textbf{10.11} &
      \textbf{10.8}                               & \textbf{3.80}                              & \textbf{5.15}          &
      \textbf{6.24}                                                                                                                        \\
      \cline{2-12}
                                                  & \multirow{2}{*}{Motifs\cite{motif}}
                                                  & baseline                                   & ~16.26~                & 18.79
                                                  & 19.69                                      & ~7.90~                 & 9.39
                                                  & 9.97                                       & 3.53                   & 5.22           &
      6.65                                                                                                                                 \\
                                                  &                                            &
      C-bias                                      & \textbf{16.61}                             & \textbf{20.38}
                                                  & \textbf{21.87}                             & \textbf{8.50}          & \textbf{9.89}  &
      \textbf{\textbf{10.41}}                     & \textbf{3.91}                              & \textbf{\textbf{5.49}} &
      \textbf{\textbf{6.99}}                                                                                                               \\
      \cline{2-12}
                                                  & \multirow{2}{*}{BGNN\cite{bgnn}}
                                                  & baseline                                   & ~26.21                 & 30.64
                                                  & 32.76                                      & 13.24                  & 15.58
                                                  & 16.58                                      & 7.82                   & 10.59          &
      12.75                                                                                                                                \\
                                                  &                                            &
      C-bias                                      & \textbf{31.30}                             &
      \textbf{\textbf{36.31}}~                    & \textbf{\textbf{38.44}}                    & \textbf{15.80}         &
      \textbf{\textbf{20.38}}                     & \textbf{\textbf{21.87}}                    & \textbf{11.63~}        &
      \textbf{\textbf{14.43}}                     & \textbf{\textbf{17.24}}                                                                \\
      \hline
      \multirow{8}{*}{no constraint}              & PCPL*                                      &
      /                                           & /                                          & 50.6
                                                  & 62.6                                       & /                      & 26.8           &
      32.8                                        & /                                          & 10.4                   & 14.4
      \\
                                                  & IMP*                                       &
      /                                           & /                                          & 9.8
                                                  & 10.5                                       & /                      & 5.8            &
      6.0                                         & /                                          & 3.8                    & 4.8
      \\
      \cline{2-12}
                                                  & \multirow{2}{*}{G-RCNN}                    &
      baseline                                    & 21.66                                      & 35.74
                                                  & 46.02                                      & 13.66                  & 20.11          &
      24.71                                       & \textbf{4.04}                              & 6.62                   & 10.11
      \\
                                                  &                                            &
      C-bias                                      & \textbf{23.31}                             & \textbf{36.38}
                                                  & \textbf{46.45}                             & \textbf{13.83}         & \textbf{20.77} &
      \textbf{26.23}                              & 4.03                                       & \textbf{6.66}          &
      \textbf{10.27}                                                                                                                       \\
      \cline{2-12}
                                                  & \multirow{2}{*}{Motifs}                    &
      baseline                                    & 23.12                                      & 35.90
                                                  & 47.72                                      & 12.90                  & 19.76          &
      25.54                                       & 4.31                                       & 7.37                   & 10.72
      \\
                                                  &                                            &
      C-bias                                      & \textbf{23.95}                             & \textbf{37.12}
                                                  & \textbf{49.67}                             & \textbf{13.59}         & \textbf{20.73} &
      \textbf{26.08}                              & \textbf{4.59}                              & \textbf{7.57}          &
      \textbf{11.55}                                                                                                                       \\
      \cline{2-12}
                                                  & \multirow{2}{*}{BGNN}                      &
      baseline                                    & 34.41                                      & 49.02
                                                  & 61.04                                      & 17.96                  & 25.29          &
      30.98                                       & 8.66                                       & 13.63                  & 17.97
      \\
                                                  &                                            &
      C-bias                                      & \textbf{36.94}                             & \textbf{51.20}
                                                  & \textbf{63.02}                             & \textbf{20.10}         & \textbf{28.01} &
      \textbf{33.65}                              & \textbf{13.57}                             & \textbf{17.07}         &
      \textbf{21.98}                                                                                                                       \\
      \hline
    \end{tabular}
  }
\end{table}
\subsection{Quantitative Results}\label{Quantitative}
\noindent\textbf{Mean Recall:}	Table \ref{tab:eval} shows comparisons between
various methods and our C-bias framework on constrained mean recall(mR@K) on VG
dataset. We notice a consistent improvement on mR@K on all baseline models.  By
plugging our framework in BGNN, our framework surpasses prior SOTA performance
in three tasks yielded by \cite{pcpl} and\cite{bgnn} (mR@100: 37.8\%, 16.6\%,
12.8\% ). It is attributed to valid local and global language representation
merging paradigms. To compare with baseline BGNN\cite{bgnn}, our framework
gains relative improvements of 17.3\% , 31.9\%	and 35.2\% on mean recall@100
metrics.  The mR@K improvement should attribute to enhanced feature powered by
language's generalization and invariability capability, which reduces noisy
propagation between nodes and edges, therefore, elevates tail relation
predictions.
In terms of Motifs and G-RCNN,	relatively minor improvements are obtained from
3 tasks over mR@k. It may caused by lack of pair-wise edge feature (e.g., union
features) utilization,	In consequence, the second paradigms can only be
applied to the last perception layer, leading to poor performance gain.
However, we find that some wrong predictions of our framework are more accurate
than annotations, which is elaborated in Section\ref{qual}.\\

\noindent\textbf{No-graph Constraint Mean Recall:}   Table \ref{tab:eval} also
reports no graph constraint mean recall(ng-mR) performance on the VG dataset,
which reveals consistent superior results on baseline models. With baseline
BGNN,ng-mR@100 increases  3.2\%, 8.6\%, 22.31\%.  Results reflects our
framework's capability of predicting correct predicates with higher confidence
when there is no graph constraint.\\

\noindent\textbf{Zero-shot Recall:} We also measure zero-shot recall (zR@K) for
evaluating the framework generalization capability, results are shown in
Table\ref{tab:eval2}. In each task, our framework still has noticeable
improvements in PredCls tasks, with relative improvement 16.37\%, 5.72\%,
33.4\% on zR@100 in 3 baselines, due to available of ground truth labels.
However, the training process shows a relatively contradictory tendency between
mR@K and zR@K performance, revealing that those
\textit{subject-predicate-object} combinations that do not occur in the
training set have a considerable proportion of high-frequency predicates.
Limited by length of paper, more  experiments would be posted on supplement
materials.

\begin{table}[!t]

  \centering
  \caption{Zero shot Recall@K evaluation in \% on VG dataset}
  \label{tab:eval2}
  \begin{tabular}{cccccccc}
    \hline
    \multirow{2}{*}{Model}        & \multirow{2}{*}{Method}       & \multicolumn{2}{c}{PredCls}
                                  & \multicolumn{2}{c}{SGCls }    &
    \multicolumn{2}{c}{SGGen }                                                                            \\
                                  &                               & ~ ~zR@50                      &
    \multicolumn{1}{l}{zR@100~ ~} & ~ ~zR@50                      & \multicolumn{1}{l}{zR@100~ ~} &
    ~ ~zR@50~                     & \multicolumn{1}{l}{zR@100~ ~}                                         \\
    \hline
    \multirow{2}{*}{G-RCNN}       & baseline                      & 10.22                         & 12.22
                                  & \textbf{6.67}                 & \textbf{7.56}                 & 0.44
                                  & 0.44                                                                  \\
                                  & C-bias                        & \textbf{12.44}                &
    \textbf{14.22}                & 4.67                          & 6.44                          &
    0.44                          & 0.44                                                                  \\
    \hline
    \multirow{2}{*}{MOTIFS}       & baseline                      & 10.22~                        & 11.70
                                  & \textbf{6.22}                 & 6.22                          & 0.00
                                  & 0.00                                                                  \\
                                  & C-bias                        & \textbf{10.44}                &
    \textbf{12.37}                & 5.33                          & \textbf{6.67}                 &
    \textbf{0.00}                 & \textbf{0.22}                                                         \\
    \hline
    \multirow{2}{*}{BGNN}         & baseline                      & 4.44                          & 5.33
                                  & 3.56                          & 3.56                          &
    \textbf{0.44}                 & 0.44                                                                  \\
                                  & C-bias                        & \textbf{6.22}                 &
    \textbf{7.11}                 & 3.56                          & \textbf{4.0}                  &
    0.00                          & \textbf{1.33}                                                         \\
    \hline
  \end{tabular}
\end{table}

\subsection{Predicate Analyze:}
\begin{figure}[t]
  \centering
  \includegraphics[width=0.96\textwidth]{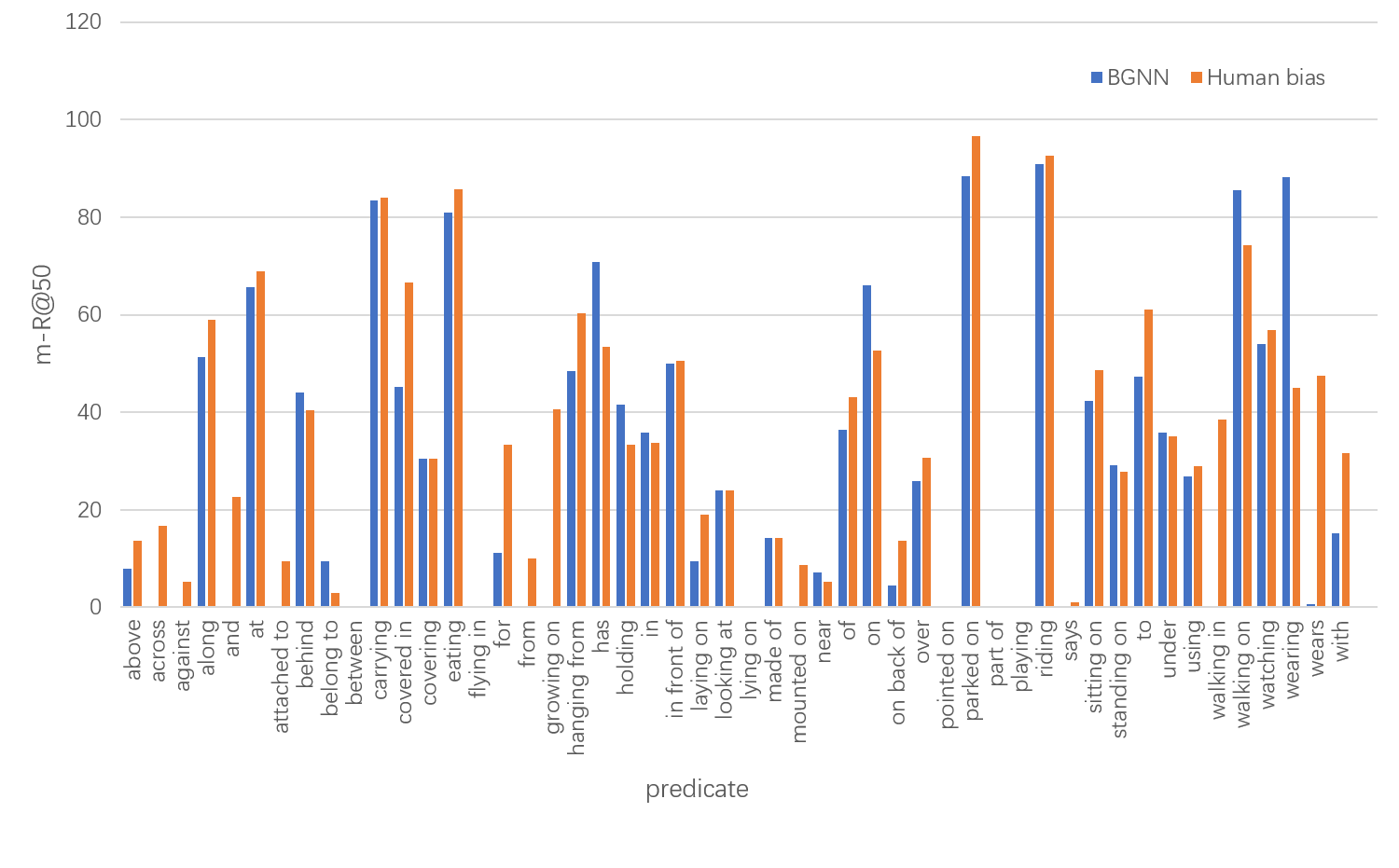}
  \caption{Comparison of mR@100 performance  on PredCls task. Orange pillars
    are BGNN, blue one is our framework. }
  \label{predicate}
\end{figure}

Shown in Figure\ref{predicate}, we present each predicate category's mR@100
performance on PredCls task. Among which yellow pillars represent Baseline
BGNN, whereas blue ones represent the proposed C-bias framework. Even compared
with such a strong baseline model, We still notice more precise predictions
that our framework gets, from which 44 of predicates are better recalled by our
framework. Besides, there are predicates that the baseline model fails to
recall (e.g.\textit{across, against, growing on, mounted on, says, walking in},
etc.) been successfully hit by the proposed C-bias framework, which supports
the claim that our framework successfully inducts relation patterns between
objects.

\subsection{Ablation Studies}
To reveal the effectiveness of each paradigm, we incrementally add them to the
baseline one by one. For simplicity, we choose BGNN as the baseline for
illustration. The results are shown in Table.\ref{tab:ab1}.\\

\noindent\textbf{Paradigm1:}We observe that adding the Experience estimation
module improves baseline with 6.83\%, 15.07\%, 32.70\% over three tasks on
mR@100 respectively, which accounts for relation distribution learned from
supervision of joint distribution. Besides, compared with the baseline that
using FREQ\cite{motif}, results prove the hypothesis that joint possibility
indeed describes real-world distribution more precisely.\\

\noindent\textbf{Paradigm2:} Furthermore, the Language map module promotes the
performance to 37.68, 19.82, 17.15. Noticed that the gap between mR@50 and
mR@100 shrinks, indicating a lower false prediction proportion in top guesses.
It is expected that channel-wised language map effectively guides union visual
feature finding intrinsic object-relation patterns, which in turn leads to the
confidence of positive-true predictions boosting.\\
\begin{table}[t]
  \centering
  \caption{Ablation studies on framework structure}
  \label{tab:ab1}
  \resizebox{\linewidth}{!}{%
    \begin{tabular}{ccccccc}
      \hline
      \multirow{2}{*}{\begin{tabular}[c]{@{}c@{}}Model\\\end{tabular}} &
      \multicolumn{2}{c}{PredCls}                                      & \multicolumn{2}{c}{SGCls}
                                                                       & \multicolumn{2}{c}{SGGen}                                    \\
                                                                       & mR@50~
                                                                       & mR@100                    & mR@50~                  & mR@100
                                                                       & mR@50~                    & mR@100                           \\
      \hline
      BGNN                                                             & 30.40~
                                                                       & 32.76                     & 14.30                   & 16.58
                                                                       & 10.70~                    & 12.75                            \\
      BGNN+EEM                                                         & 32.75
                                                                       & 35.00                     & 17.77                   & 19.08
                                                                       & 14.54                     & 16.92                            \\
      BGNN+EEM+LMM                                                     & 35.99
                                                                       & 37.68                     & 18.71                   & 19.82
                                                                       & \textbf{15.10}            & 17.15                            \\
      BGNN+EEM+LMM+SEM                                                 &
      \textbf{36.31}                                                   & \textbf{\textbf{38.44}}   & \textbf{20.38}          &
      \textbf{\textbf{21.87}}                                          & 14.43                     & \textbf{\textbf{17.24}}          \\
      \hline
      \multicolumn{1}{l}{}                                             &
      \multicolumn{1}{l}{}                                             & \multicolumn{1}{l}{}      & \multicolumn{1}{l}{}    &
      \multicolumn{1}{l}{}                                             & \multicolumn{1}{l}{}      & \multicolumn{1}{l}{}
    \end{tabular}
  }
\end{table}

\noindent\textbf{Paradigm3:} By adding the Scene extractor module, the
performances further reach 38.44, 20.87, and 17.24. Through SEM brings a
relatively tiny improvement than other modules, it meets our anticipation that
global language context mainly serves as the assistance when a label is
ambiguous.

\subsection{Model Size and Runtime:}
We present several model parameters and inference time in
Table\ref{tab:runtime}. We choose BGNN and MOTIFS as baselines, C-bias and
Unbiased\cite{unbias} as a plug-in model. Though Unbiased is claimed as a
inference  method, our framework still shows low computational requirements
than Unbiased.	For C-bias, parameters of model merely increased 5.33\% and
6.17\% in two baselines, the performance of mean recall@100 increases 24.88\%
and 8\%, Compare with unbiased, our framework size is smaller but perform
better. This advantage comes from label input of 3 paradigms, which is
low-dimensional than image.
\begin{table}[t]
  \centering
  \caption{Compariation of model size and runtime }
  \label{tab:runtime}
  \resizebox{\linewidth}{!}{%
    \begin{tabular}{cccccc}
      \toprule
      Model              & ~ ~BGNN~ ~        & ~ ~BGNN+Ours~ ~ & ~ ~BGNN+Unbiased~ ~ & ~
      ~MOTIFS~ ~         & ~ ~MOTIFS+Ours~ ~                                                 \\
      \midrule
      Params             & 341.9M            & 360.13M         & 365.746M            &
      367.17M            & 389.82M                                                           \\
      Relative increase  & /                 & 5.33\%          & 6.98\%              & /
                         & 6.17\%                                                            \\
      Inference Time(ms) & 497.5             & 538.1           & 539.2               & 420.4
                         & 450.9                                                             \\
      Mean recall@100    & 32.76             & 38.44           & 34.89               & 6.65
                         & 6.99                                                              \\
      \bottomrule
    \end{tabular}
  }
\end{table}
\begin{figure}[t]
  \centering

  \includegraphics[width=0.98\textwidth]{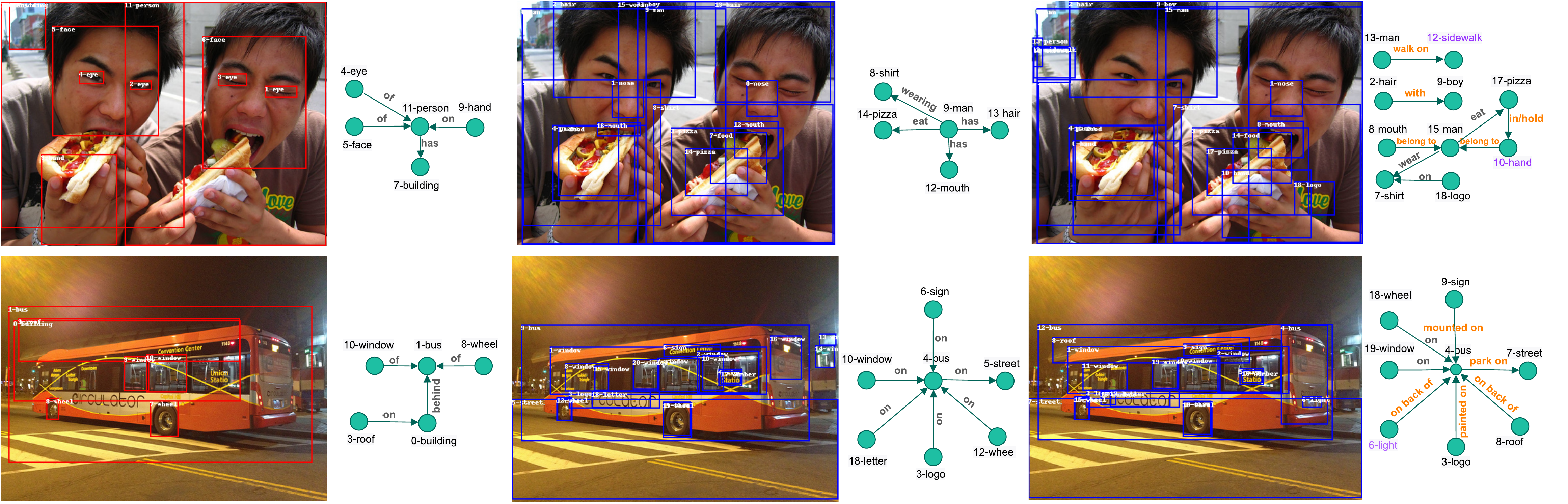}



  \caption{Visualization Results: we present scene graphs generated by
    annotations, baseline BGNN, ours in three columns, respectively. The relations
    and entities that neither occur in annotations nor baseline are marked with
    purple and orange.}
  \label{visual}
\end{figure}

\subsection{Qualitative Studies}
\label{qual}
Beyond numbers, we emphasize the intuitive evaluation because trivial and
uncompleted ground truth has traumatized the authority of metrics.    Some
inspiring findings that we discovered from wrong results of SGGen task in
Figure\ref{visual} reveal several advantages that the C-bias framework has over
baseline and annotations, including:

\begin{enumerate}
  \item \textbf{Multi-scaled:} An intuitive sensation is that our results
        generate more relations widespread in the whole scene. Compared to baseline
        results which are incapable of detecting relations with no overlaps, our C-bias
        framework successfully recognizes distanced entities' correlations. In Figure
        \ref{visual}, the last column of first image shows our framework can notice
        $<$\textit{man walk on sidewalk}$>$ in a small area of the background.

  \item \textbf{Accurate and Informative:} The baseline predicts
        multitudinous less meaningful categories \textit{"on, in, of"}. We speculate
        that it is caused by massive repeated annotations full in annotations (ground
        truth graph in the first row of Figure \ref{visual}), which indeed deteriorate
        model learning. Intuitively, we find that our proposed framework can mine
        meaningful but inconspicuous correlations regardless of trivial annotations.
        Take last row of Figure\ref{visual} for example, \textit{"letter"} are neither
        part of annotations or baseline results, and  $<$\textit{logo on bus}$>$ in
        annotations is not informative. Inspiringly, in the C-bias framework, relation
        triblets  $<$\textit{letter painted on bus}$>$ and  $<$\textit{logo mounted on
          bus}$>$ are more accurate.
\end{enumerate}

\section{Conclusion}
In this paper, we propose a novel cognitive bias framework   in SGG task to
properly mining fickle relation patterns by incorporating local, global context
with help of linguistics information. We simulate human cognitive bias
paradigms to acquire better relation perception capability, and address crucial
role of using label-derivative linguistic features as biases upon visual
message propagation. The result shows superiority over previous works.

\clearpage
\bibliography{egbib}
\bibliographystyle{splncs04}

\end{document}